# A pathway-based kernel boosting method for sample classification using genomic data

LI ZENG

*Department of Biostatistics, Yale University, New Haven, CT 06511, USA*

ZHAOLONG YU

*Interdepartmental Program in Computational Biology and Bioinformatics, Yale University, New Haven, CT 06511, USA*

HONGYU ZHAO∗

*Department of Biostatistics, Yale University, New Haven, CT 06511, USA*

hongyu.zhao@yale.edu

SUMMARY

The analysis of cancer genomic data has long suffered "the curse of dimensionality". Sample sizes for most cancer genomic studies are a few hundreds at most while there are tens of thousands of genomic features studied. Various methods have been proposed to leverage prior biological knowledge, such as pathways, to more effectively analyze cancer genomic data. Most of the methods focus on testing marginal significance of the associations between pathways and clinical phenotypes. They can identify relevant pathways, but do not involve predictive modeling. In this article, we propose a Pathway-based Kernel Boosting (PKB) method for integrating gene pathway information for sample classification, where we use kernel functions calculated from each pathway as base learners and learn the weights through iterative optimization of the classifica-tion loss function. We apply PKB and several competing methods to three cancer studies with pathological and clinical information, including tumor grade, stage, tumor sites, and metastasis status. Our results show that PKB outperforms other methods, and identifies pathways relevant to the outcome variables.

∗To whom correspondence should be addressed.

## 1. Introduction

High-throughput genomic technologies have enabled cancer researchers to study the associations between genes and clinical phenotypes of interest. A large number of cancer genomic data sets have been collected with both genomic and clinical information from the patients. The analyses of these data have yielded valuable insights on cancer mechanisms, subtypes, prognosis, and treatment response.

Although many methods have been developed to identify genes informative of clinical phenotypes and build prediction models from these data, it is often difficult to interpret the results with single-gene focused approaches, as one gene is often involved in multiple biological processes, and the results are not robust when the signals from individual genes are weak. As a result, pathway-based methods have gained much popularity (e.g. Subramanian *and others* (2005)). A pathway can be considered as a set of genes that are involved in the same biological process or molecular function. It has been shown that gene-gene interactions may have stronger effects on phenotypes when the genes belong to the same pathways or regulatory networks (Carlson *and others*, 2004). There are many pathway databases available, such as the Kyoto Encyclopedia of Genes and Genomes (Kanehisa and Goto, 2000), the Pathway Interaction Database (Schaefer *and others*, 2008), and Biocarta (Nishimura, 2001). By utilizing pathway information, researchers may aggregate weak signals from the same pathway to identify relevant pathways with better power and interpretability. Many pathway-based methods, such as GSEA (Subramanian *and others*, 2005), LSKM (Liu *and others*, 2007), and SKAT (Wu *and others*, 2011), focus on testing the significance of pathways. These methods consider each pathway separately and evaluate statistical significance for its relevance to the phenotype. In other words, these methods study each pathway separately without considering the effects of other pathways.

Given that many pathways likely contribute to the onset and progression of a disease, it is of interest to study the contribution of a specific pathway to phenotypes conditional on the effects of other pathways. This is usually achieved by regression models. Wei and Li (2007) and Luan and Li (2007) proposed two similar models, Nonparametric Pathway-based Regression (NPR) and Group Additive Regression (GAR). Both models employ a boosting framework, construct base learners from individual pathways, and perform prediction through additive models. Due to the additivity at the pathway level, these models only considered interactions among genes within the same pathway, but not across pathways. Since our proposed method is motivated by the above two models, more details of these methods will be described in Section 2. In genomics data analysis, multiple kernel methods (Gönen and Margolin, 2014; Aiolli and Donini, 2015) are also commonly used when predictors have group structures. In these methods, one kernel is assigned to each pathway and a meta-kernel is computed as a weighted sum of the individual kernels. The kernel weights are estimated through optimization, and can be considered as a measure of pathway importance. Although multiple kernel methods are not designed specifically for integrating pathway information, they have been successful in genomic data analysis (Costello *and others*, 2014).

In this paper, we propose a Pathway-based Kernel Boosting (PKB) method for sample classification. In our boosting framework, we use the second order approximation of the loss function instead of the first order approximation used in the usual gradient descent boosting method, which allows for deeper descent at each step. We introduce two types of regularizations ($L_1$ and $L_2$) for selection of base learners in each iteration, and propose algorithms for solving the regularized problems. In Section 3, we conduct simulation studies to evaluate the performance of PKB, along with four other competing methods. In Section 4, we apply PKB to three cancer genomic data sets, where we use gene expression data to predict several patient phenotypes, including tumor grade, stage, tumor site, and metastasis status.

## 2. METHODS

Suppose our observed data is collected from $N$ subjects. For subject $i$, we use a $p$ dimensional vector $\mathbf{x}_i = (x_{i1}, x_{i2}, \ldots, x_{ip})$ to denote the normalized gene expression profile, and $y_i \in \{1, -1\}$ to denote its class label. Similarly, the gene expression levels of a given pathway $m$ with $p_m$ genes can be represented by $\mathbf{x}_i^{(m)} = (\mathbf{x}_{i1}^{(m)}, \mathbf{x}_{i2}^{(m)}, \ldots, \mathbf{x}_{ip_m}^{(m)})$, which is a sub-vector of $\mathbf{x}_i$.

The log loss function is commonly used in binary classifications with the following form:

$$l(y, F(\mathbf{x})) = \log(1 + e^{-yF(\mathbf{x})}),$$

and is minimized by

$$F^*(\mathbf{x}) = \log \frac{p(y=1|\mathbf{x})}{p(y=-1|\mathbf{x})},$$

which is exactly the log odds function. Thus the sign of an estimated $F(\mathbf{x})$ can be used to classify sample $\mathbf{x}$ as 1 or -1. Since genes within the same pathway likely have much stronger interactions than genes in different pathways, in our pathway-based model setting, we assume additive effects across pathways, and focus on capturing gene interactions within pathways:

$$F(\mathbf{x}) = \sum_{m=1}^{M} H_m(\mathbf{x}^{(m)}),$$

where each $H_m$ is a nonlinear function that only depends on expressions of genes in the $m$th pathway, and summarizes its contribution to the log odds function. Due to the additive nature of this model, it only captures gene interactions within each pathway, but not across pathways.

Two existing methods, NPR (Wei and Li, 2007) and GAR (Luan and Li, 2007), employed the Gradient Descent Boosting (GDB) framework (Friedman, 2001) to estimate the functional form of $F(\mathbf{x})$ nonparametrically. GDB can be considered as a functional gradient descent algorithm to minimize the empirical loss function, where in each descent iteration, an increment function that best aligns with the negative gradient of the loss function (evaluated at each sample point) is selected from a space of base learners and then added to the target function $F(\mathbf{x})$. NPR and GAR extended GDB to be pathway-based by applying the descent step to each pathway separately, and selecting the base learner from the pathway that provides the best fit to the negative gradient.

NPR and GAR differ in how they construct base learners from each pathway: NPR uses regression trees, and GAR uses linear models. Due to the linearity assumption of GAR, it lacks the ability to capture complex interactions among genes in the same pathways. Using regression tree as base learners enables NPR to model interactions, however, there is no regularization in the gradient descent step, which can lead to selection bias that prefers large pathways.

Motivated by NPR and GAR, we propose the PKB model, where we employ kernel functions as base learners, optimize loss function with second order approximation (Friedman *and others*, 2000) which gives Newton-like descent speed, and also incorporates regularization in selection of pathways in each boosting iteration.

### 2.1 *PKB model*

Kernel methods have been applied to a variety of statistical problems, including classification (Guyon *and others*, 2002), regression (Drucker *and others*, 1997), dimension reduction (Fukumizu *and others*, 2009), and others. Results from theories of Reproducing Kernel Hilbert Space (Friedman *and others*, 2001) have shown that kernel functions can capture complex interactions among features. For pathway $m$, we construct a kernel-based function space as the space for base

learners

$$\mathcal{G}_m = \{g(\mathbf{x}) = \sum_{i=1}^{N} K_m(\mathbf{x}_i^{(m)}, \mathbf{x}^{(m)})\beta_i + c : \beta_1, \beta_2, \ldots, \beta_N, c \in R\},$$

where $K_m(\cdot, \cdot)$ is a kernel function that defines similarity between two samples only using genes in the $m$th pathway. The overall base learner space is the union of the spaces constructed from each pathway alone: $\mathcal{G} = \cup_{m=1}^{M} \mathcal{G}_m$.

Estimation of the target function $F(\mathbf{x})$ is obtained through iterative minimization of the empirical loss function evaluated at the observed data. The empirical loss is defined as

$$L(\mathbf{y}, \mathbf{F}) = \frac{1}{N} \sum_{i=1}^{N} l(y_i, F(\mathbf{x}_i)),$$

where $\mathbf{F} = (F(\mathbf{x}_1), F(\mathbf{x}_2), \ldots, F(\mathbf{x}_N))$. In the rest of this article, we will use bold font of a function to represent the vector of the function evaluated at the observed $\mathbf{x}_i$'s. Assume that at iteration $t$, the estimated target function is $F_t(\mathbf{x})$. In the next iteration, we aim to find the best increment function $f \in \mathcal{G}$, and add it to $F_t(\mathbf{x})$. Expanding the empirical loss at $\mathbf{F}_t$ to the second order, we can get the following approximation

$$L_{\text{approx}}(\mathbf{y}, \mathbf{F}_t + \mathbf{f}) = L(\mathbf{y}, \mathbf{F}_t) + \frac{1}{N} \sum_{i=1}^{N} [h_{t,i} f(\mathbf{x}_i) + \frac{1}{2} q_{t,i} f(\mathbf{x}_i)^2], \quad (2.1)$$

where

$$h_{t,i} = \frac{\partial L(\mathbf{y}, \mathbf{F}_t)}{\partial F_t(\mathbf{x}_i)} = -\frac{y_i}{1 + e^{y_i F_t(\mathbf{x}_i)}},$$

$$q_{t,i} = \frac{\partial^2 L(\mathbf{y}, \mathbf{F}_t)}{\partial F_t(\mathbf{x}_i)^2} = \frac{e^{y_i F_t(\mathbf{x}_i)}}{(1 + e^{y_i F_t(\mathbf{x}_i)})^2}$$

are the first order and second order derivatives with respect to each $F_t(\mathbf{x}_i)$, respectively. We propose a regularized loss function that incorporates both the approximated loss and a penalty on the complexity of $f$:

$$L_R(\mathbf{f}) = L_{\text{approx}}(\mathbf{y}, \mathbf{F}_t + \mathbf{f}) + \Omega(f) \quad (2.2)$$

$$= \frac{1}{N} \sum_{i=1}^{N} \frac{q_{i,t}}{2} (\frac{h_{i,t}}{q_{i,t}} + f(\mathbf{x}_i))^2 + \Omega(f) + C(\mathbf{y}, \mathbf{F}_t), \quad (2.3)$$

*PKB: Pathway-based Kernel Boosting* 7Table 1. An overview of the PKB algorithm

1. Initialize target function as an optimal constant:

$$F_0(\mathbf{x}) = \arg\min_{r \in R} \frac{1}{n} \sum_{i=1}^{N} l(y_i, r)$$

For t from 0 to T-1 (maximum number of iterations) do:

2. calculate first and second derivatives:

$$h_{t,i} = -\frac{y_i}{1 + e^{y_i F_t(\mathbf{x}_i)}}, q_{t,i} = \frac{e^{y_i F_t(\mathbf{x}_i)}}{(1 + e^{y_i F_t(\mathbf{x}_i)})^2}$$

3. optimize the regularized loss function in the base learner space:

$$\hat{f} = \arg\min_{f \in \mathcal{G}} L_R(\mathbf{f})$$

4. find the step length with the steepest descent:

$$\hat{d} = \arg\min_{d \in R^+} L(\mathbf{y}, \mathbf{F}_t + d\hat{f})$$

5. update the target function:

$$F_{t+1}(\mathbf{x}) = F_t(\mathbf{x}) + \nu \hat{d} \hat{f}(\mathbf{x})$$

End For
return $F_T(\mathbf{x})$

where $\Omega(\cdot)$ is the penalty function. Since $f \in \mathcal{G}$ is a linear combination of kernel functions calculated from a specific pathway, the norm of the combination coefficients can be used to define $\Omega(\cdot)$. We consider both $L_1$ and $L_2$ norm penalties, and solutions regarding each penalty option are presented in sections 2.1.1 and 2.1.2, respectively. $C(\mathbf{y}, \mathbf{F}_t)$ is a constant term with respect to $f$. Therefore, we only use the first two terms of equation (2.3) as the working loss function in our algorithms. We will also drop $C(\mathbf{y}, \mathbf{F}_t)$ in the expression of $L_R(\mathbf{f})$ in the following sections for brevity. Such a penalized boosting step has been employed in several methods (e.g. Johnson and Zhang (2014)). Intuitively, the regularized loss function would prefer simple solutions that also fit the observed data well, which usually leads to more robust models.

We then optimize the regularized loss for the best increment direction

$$\hat{f} = \arg\min_{f \in \mathcal{G}} L_R(\mathbf{f}).$$

Given the direction, we find the deepest descent step length by minimizing over the original loss function

$$\hat{d} = \arg\min_{d \in R^+} L(\mathbf{y}, \mathbf{F}_t + \hat{d}\hat{f}),$$

and update the target function to $F_{t+1}(\mathbf{x}) = F_t(\mathbf{x}) + \nu \hat{d}\hat{f}$, where $\nu$ is a learning rate parameter. The above fitting procedure is repeated until a certain pre-specified number of iterations is reached. A complete procedure of the PKB algorithm is illustrated in Table 1.

2.1.1 *$L_1$ penalized boosting*

The core step of PKB is the optimization of the regularized loss function (see step 3 of Table 1). Note that $\mathcal{G}$ is the union of the pathway-based learner spaces, thus

$$\hat{f} = \arg\min_{\mathcal{G}} L_R(\mathbf{f})$$
$$= \arg\min_{\hat{f}_m} \{L_R(\hat{f}_m) : \hat{f}_m = \arg\min_{f \in \mathcal{G}_m} L_R(\mathbf{f}), m = 1, 2, \ldots, M\}.$$

To solve for $\hat{f}$, it is sufficient that we obtain the optimal $\hat{f}_m$ in each pathway-based subspace. Due to the way we construct the subspaces, in a given pathway $m$, $f$ takes a parametric form as linear combination of the corresponding kernel functions. This helps us further reduce the optimization problem to

$$\min_{f \in \mathcal{G}_m} L_R(\mathbf{f}) = \min_{\beta,c} \frac{1}{N} \sum_{i=1}^{N} \frac{q_{i,t}}{2} \left( \frac{h_{i,t}}{q_{i,t}} + K_{m,i}^T \beta + c \right)^2 + \Omega(f) \qquad (2.4)$$

$$= \min_{\beta,c} \frac{1}{N} (\eta_t + K_m\beta + 1_N c)^T W_t (\eta_t + K_m\beta + 1_N c) + \Omega(f), \qquad (2.5)$$

where

$$\eta_t = \left( \frac{h_{1,t}}{q_{1,t}}, \frac{h_{2,t}}{q_{2,t}}, \ldots, \frac{h_{N,t}}{q_{N,t}} \right)^T,$$
$$W_t = \text{diag}\left( \frac{q_{1,t}}{2}, \frac{q_{2,t}}{2}, \ldots, \frac{q_{N,t}}{2} \right),$$
$$K_m = \left[ K_m(\mathbf{x}_i^{(m)}, \mathbf{x}_j^{(m)}) \right]_{i,j=1,2,\ldots,N}.$$

$K_{m,i}$ is the $i$th column of kernel matrix $K_m$, and $1_N$ is an $N$ by 1 vector of 1's. We use the $L_1$ norm $\Omega(f) = \lambda\|\beta\|_1$, as the penalty term, where $\lambda$ is a tuning parameter adjusting the amount of penalty we impose on model complexity. We also prove that after certain transformations, the optimization can be converted to a LASSO problem without intercept

$$\min_{\beta} \frac{1}{N}\|\tilde{\eta} + \tilde{K}_m\beta\|_2^2 + \lambda\|\beta\|_1, \tag{2.6}$$

where

$$\tilde{\eta} = W_t^{\frac{1}{2}}\left[I_N - \frac{1_N 1_N^T W_t}{tr(W_t)}\right]\eta_t$$

$$\tilde{K}_m = W_t^{\frac{1}{2}}\left[I_N - \frac{1_N 1_N^T W_t}{tr(W_t)}\right]K_m.$$

Therefore, $\beta$ can be efficiently estimated using existing LASSO solvers. The proof of the equivalence between the two problems is provided in Section S1 of the supplementary materials.

### 2.1.2 $L_2$ penalized boosting

In the $L_2$ penalized boosting, we replace $\Omega(f)$ in the objective function of (2.5) with $\lambda\|\beta\|_2^2$. Following the same transformation as in section 2.1.1, the objective can also be converted to a standard Ridge Regression (see S1 of supplementary material)

$$\min_{\beta} \frac{1}{N}\|\tilde{\eta} + \tilde{K}_m\beta\|_2^2 + \lambda\|\beta\|_2^2, \tag{2.7}$$

which allows closed form solution

$$\hat{\beta} = -(\tilde{K}_m^T\tilde{K}_m + N\lambda I_N)^{-1}\tilde{K}_m^T\tilde{\eta}.$$

Both the $L_1$ and $L_2$ boosting algorithms require the specification of the penalty parameter $\lambda$, which controls step length (the norm of fitted $\beta$) in each iteration, and additionally controls solution sparsity in the $L_1$ case. Feasible choices of $\lambda$ might be different for different scenarios, depending on the input data and also the choice of the kernel. Either too small or too large $\lambda$ values would lead to big leaps or slow descent speed. Under the $L_1$ penalty, poor choices of $\lambda$ can even result in all-zero $\beta$, which makes no change to the target function. Therefore, we also incorporate an optional automated procedure to choose the value of $\lambda$ in PKB. Computational details of the procedure are provided in Section S2 of the supplementary materials. We recommend the use of the automated procedure to calculate a feasible $\lambda$, and try a range of values around it (e.g. the calculated value multiplies $1/25, 1/5, 1, 5, 25$) for improved performance.

Lastly, the final target function at iteration $T$ can be written as

$$F_T(\mathbf{x}) = \sum_{m=1}^{M} \sum_{i=1}^{N} K_m(\mathbf{x}_i^{(m)}, \mathbf{x}^{(m)}) \beta_i^{(m)} + C,$$

where $\beta^{(m)} = (\beta_1^{(m)}, \beta_2^{(m)}, \ldots, \beta_N^{(m)})$ are the combination coefficients of kernel functions from pathway $m$. We use $\|\beta^{(m)}\|_2$ as a measure of importance (or weight) in the target function. It is obvious that only the pathways that are selected at least once in the boosting procedure will have non-zero weights. Because $F_T(\mathbf{x})$ is an estimation of the log odds function, $\text{sign}[F_T(\mathbf{x})]$ is used as the classification rule to assign $\mathbf{x}$ to 1 or -1.

## 3. Simulation studies

We use simulation studies to assess the performance of PKB. We consider the following three underlying truth models:

- Model 1:
$$F(\mathbf{x}) = 2x_1^{(1)} + 3x_2^{(1)} + \exp(0.8x_1^{(2)} + 0.8x_2^{(2)}) + 4x_1^{(3)} x_2^{(3)}$$

- Model 2:
$$F(\mathbf{x}) = 4\sin(x_1^{(1)} + x_2^{(1)}) + 3|x_1^{(2)} - x_2^{(2)}| + 2x_1^{(3)^2} - 2x_2^{(3)^2}$$

- Model 3:
$$F(\mathbf{x}) = 2 \sum_{m=1}^{10} \|\mathbf{x}^{(m)}\|_2$$

where $F(\mathbf{x})$ is the true log odds function, and $x_i^{(m)}$ represents the expression level of the $i$th gene in the $m$th pathway. We include different functional forms of pathway effects in $F(\mathbf{x})$, including linear, exponential, polynomial and others. In models 1 and 2, only two genes in each of the first three pathways are informative to sample classes; in model 3, only genes in the first ten pathways are informative. We generated a total of six datasets, two for each model, with different numbers of irrelevant pathways (M= 50 and 150) corresponding to different noise levels. We set the size of pathways to 5 and sample size to 900 in all simulations. Gene expression data ($\mathbf{x}_i$'s) were generated following standard normal distribution. We then calculated the log odds $F(\mathbf{x}_i)$ for each sample, and use the median-centered $F(\mathbf{x_i})$ values to generate corresponding binary outcomes $y_i \in \{-1, 1\}^*$.

We divided the generated datasets into three folds, and each time used two folds as training data and the other fold as testing data. The number of maximum iterations $T$ is important to PKB, as using a large $T$ will likely induce overfitting on training data and poor prediction on testing data. Therefore, we used an additional layer of cross validation within the training data to select $T$. We further divided the training data into three folds, and each time trained the PKB model using two folds while monitoring the loss function on the other fold at every iteration. Eventually, we identified the iteration number $T^*$ with the minimum averaged loss, and applied PKB to the whole training dataset up to $T^*$ iterations.

We first evaluated the ability of PKB to correctly identify relevant pathways. For each simulation scenario, we calculated the average optimal weights across different cross validation runs, and the results are shown in Figure 1, where the X-axis represents different pathways, and the length of bars above them represents corresponding weights in the prediction functions. Note that for the underlying Model 1 and Model 2, only the first three pathways were relevant to the outcome, and in Model 3, the first ten pathways were relevant. In all the cases, PKB successfully

---

*We use the median-centered $F(\mathbf{x}_i)$ values to generate outcome, so that the proportions of 1's and -1's are approximately 50%.

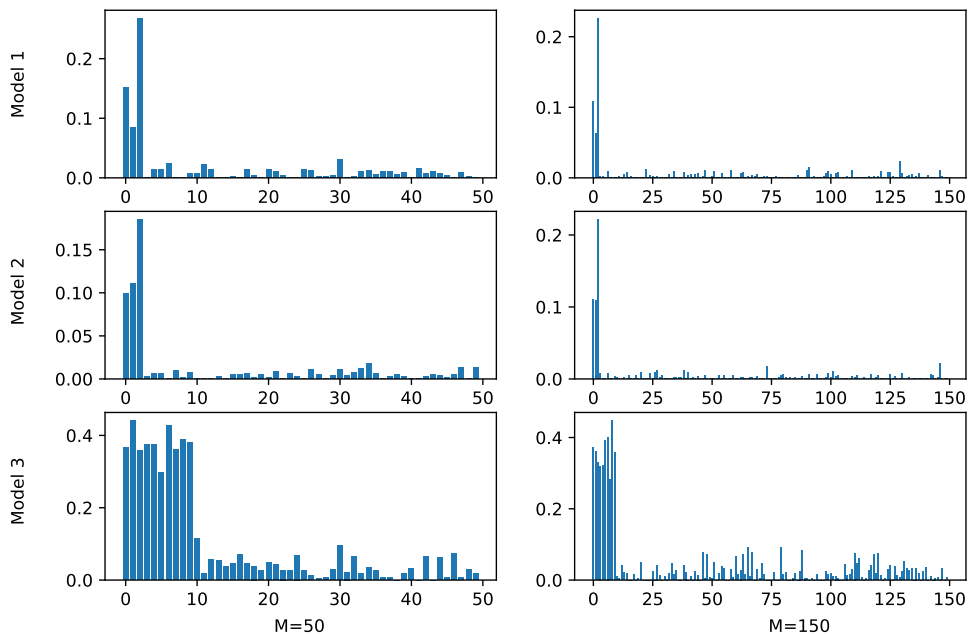

Fig. 1. Estimated pathway weights by PKB in simulation studies. The X-axis represents pathways, and the Y-axis represents estimated weights. Based on the simulation settings, the first three pathways are relevant in Models 1 and 2, and the first ten pathways are relevant in Model 3. $M$ represents the number of simulated pathways.

assigned the largest weights to relevant pathways. Since PKB is an iterative approach, at some iterations, certain pathways irrelevant to the outcome may be selected by chance and added to the prediction function. This explains the non-zero weights of the irrelevant pathways, and their values are clearly smaller than those of relevant pathways.

We also applied several commonly used methods to the simulated datasets, and compared their prediction accuracy with PKB. These methods included both non-pathway-based methods: Random Forest (Breiman, 2001) and SVM (Guyon and others, 2002), and pathway-based methods: NPR (Wei and Li, 2007) and EasyMKL (Aiolli and Donini, 2015). We used the same three-fold split of the data, as we used when applying PKB, to perform cross-validations for each competing method. The average prediction performance of the methods is summarized in

PKB: Pathway-based Kernel Boosting 13

Table 2. It can be seen that the pathway-based methods generally performed better than the non-pathway-based methods in all simulated scenarios. Among the pathway-based methods, the one that utilized kernels (EasyMKL) had comparable performances with the tree-based NPR method in Models 1 and 2, but had clearly superior performance in Model 3. This was likely due to the functional form of the log odds function $F(\mathbf{x})$ of Model 3. Note that genes in relevant pathways were involved in $F(\mathbf{x})$ in terms of their $L_2$ norms, which is hard to approximate by regression tree functions, but can be well captured using kernel methods. In all scenarios, the best performance was achieved by one of the PKB methods. In four out of six scenarios, the PKB-$L_2$ method produced the smallest prediction errors, while in the other two scenarios, PKB-$L_1$ was slightly better. Although PKB-$L_1$ and PKB-$L_2$ had similar performances, PKB-$L_1$ was usually computationally faster, because in the optimization step of each iteration, the $L_1$ algorithm only looked for sparse solution of $\beta$'s, which can be done more efficiently than PKB-$L_2$, which involves matrix inverse.

| Method | Model 1 | | Model 2 | | Model 3 | |
|---|---|---|---|---|---|---|
| | 50 | 150 | 50 | 150 | 50 | 150 |
| PKB-$L_1$ | **0.151** | 0.196 | **0.198** | 0.189 | 0.179 | 0.21 |
| PKB-$L_2$ | 0.158 | **0.185** | 0.201 | **0.183** | **0.157** | **0.173** |
| Random Forest | 0.305 | 0.331 | 0.290 | 0.328 | 0.341 | 0.400 |
| SVM | 0.353 | 0.431 | 0.412 | 0.476 | 0.431 | 0.492 |
| NPR | 0.271 | 0.321 | 0.299 | 0.317 | 0.479 | 0.440 |
| EasyMKL | 0.253 | 0.284 | 0.268 | 0.330 | 0.212 | 0.300 |

Table 2. Classification error rate from PKB and competing methods in simulation studies. The numbers below each model represent the number of pathways simulated in the data sets.

## 4. Real data applications

We applied PKB to gene expression profiles to predict clinical features in three cancer studies, including breast cancer, melanoma, and glioma. The clinical variables we considered included tumor grade, tumor site, and metastasis status, which were all of great importance to cancer.

We used three commonly used pathway databases: KEGG, Biocarta, and Gene Ontology (GO) Biological Process pathways. These databases provide lists of pathways with emphasis on different biological aspects, including molecular interactions and involvement in biological processes. The number of pathways from these databases ranges from 200 to 4,000. There is considerable overlap between pathways. To eliminate redundant information and control the overlap between pathways, we applied a preprocessing step to the databases with details provided in the supplementary material Section S4.2.

Similar to the simulation studies, we compared the performances from different methods based on three fold cross validations following the same procedure as elaborated in Section 3. Most of the methods we considered have tuning parameters. We searched through different parameter configurations, and reported the best result from cross-validation for each method. More details of the data sets and the implementations can be found in the supplementary material Section S4. Table 3 shows the classification error rates from all methods. The numbers in bold are the optimal error rates for each column separately. In four out of five classifications, PKB was the best method (usually with the $L_1$ and $L_2$ methods being the top two). In the other case (melanoma, stage), NPR yielded the best results, with the PKB methods still ranking second and third.

We provide more detailed introductions to the data sets and clinical variables, and interpretations of results by PKB in the following. For brevity of the article, we focus on presenting results for three outcomes, one from each data set, and leave the other two in the supplementary materials (Section S4.4).

### 4.1 *Breast Cancer*

Metabric is a breast cancer study that involved more than 2,000 patients with primary breast tumors (Pereira and others, 2016). The data set provides copy number aberration, gene expression, mutation, and long-term clinical follow-up information. We are interested in the clinical variable

*PKB: Pathway-based Kernel Boosting* 15| Method | Data sets | | | | |
|---|---|---|---|---|---|
| | Metabric (grade) | Glioma (grade) | Glioma (site) | Melanoma (stage) | Melanoma (met) |
| PKB-$L_1$ | **0.274** | **0.283** | 0.168 | 0.304 | **0.081** |
| PKB-$L_2$ | 0.304 | **0.283** | **0.154** | 0.307 | 0.083 |
| Random Forest | 0.306 | 0.302 | 0.306 | 0.320 | 0.136 |
| SVM | 0.285 | 0.292 | 0.185 | 0.314 | 0.083 |
| NPR | 0.306 | 0.298 | 0.197 | **0.282** | 0.110 |
| EasyMKL | 0.297 | 0.302 | 0.291 | 0.314 | 0.100 |

Table 3. Classification error rates on real data. The names in the parenthesis of each data set are the variables used as classification outcome. The best error rates are highlighted with bold font for each column.

of tumor grade, which measures the abnormality of the tumor cells compared to normal cells under a microscope. It takes a value of 1, 2, or 3. Higher grade indicates more abnormality and higher risk of rapid tumor proliferation. Since grade 1 contained the fewest samples, we pooled it together with grade 2 as one class, and treated grade 3 as the other class.

We then applied PKB to samples in subtype Lum B, where the sample sizes for the two classes were most balanced (259 Grade 3 patients; 211 Grade 1,2 patients). For input gene expression data, we used the normalized mRNA expression (microarray) data for 24,368 genes provided in the data set. The model using GO Biological Process pathways and radial basis function (rbf) kernel yielded the best performance (error rate 27.4%). To obtain the pathways most relevant to tumor grade, we calculated the average pathway weights from the cross validation, and sorted them from highest to lowest. Top fifteen pathways with the highest weights are presented in the first columns of Table 4.

Among all pathways, the cell aggregation and sequestering of metal ion are the top two pathways in terms of the estimated pathway weights. Previous research has shown that cell aggregation contributes to the inhibition of cell death and anoikis-resistance, thereby promoting tumor cell proliferation. Genes in the cell aggregation pathway include TGFB2, MAPK14, FGF4, and FGF6, which play important roles in the regulation of cell differentiation and fate (Zhang *and others*, 2010). Moreover, the majority of genes in the sequestering of metal ion pathway encode calcium-binding proteins, which regulates calcium level and different cell signaling pathways relevant to tumorigenesis and progression (Monteith and others, 2007). Among these genes, S100A8 and S100A9 have been identified as novel diagnostic markers of human cancer (Hermani and others, 2005). The results suggest that PKB has identified pathways that are likely relevant to breast cancer grade.

### 4.2 Lower Grade Glioma

Glioma is a type of cancer developed in the glial cells in brain. As glioma tumor grows, it compresses normal brain tissue and can lead to disabling or fatal results. We applied our method to a lower grade glioma data set from TCGA, where only grade 2 and 3 samples were collected[*] (TCGA and others, 2015). After removal of missing values, the numbers of patients in the cohort with grades 2 and 3 tumors were 248 and 265, respectively. We used grade as the outcome variable to be classified, and applied PKB with different parameter configurations. After cross validation, PKB using the third order polynomial (poly3) kernel and the GO Biological Process pathways yielded an error rate of 28.3%, which was the smallest among all methods. The top fifteen pathways selected in the model are listed in the second column of Table 4.

The estimated pathway weights indicate that the cell adhesion pathway and the neuropeptide signaling pathway have the strongest association with glioma grade. Genes in the cell adhesion pathways generally govern the activities of cell adhesion molecules. Turning off the expression of cell-cell adhesion molecules is one of the hallmarks of tumor cells, by which tumor cells can inhibit antigrowth signals and promote proliferation. Previous studies have shown that deletion of carcinoembryonic antigen-related cell adhesion molecule 1 (CEACAM1) gene can contribute to cancer progression (Leung and others, 2006). Cell Adhesion Molecule 1 (CADM1), CADM2, CADM3, and CADM4, serve as tumor suppressors and can inhibit cancer cell proliferation and induce apoptosis. Neuropeptide signaling pathway has also been implicated in tumor growth and progression. Neuropeptide Y is highly relevant to tumor cell proliferation and survival. Two NPY receptors, Y2R and Y5R, are also members of the neuropeptide signaling pathway. They are considered as important stimulatory mediators in tumor cell proliferation (Tilan and Kitlinska, 2016).

---

[*]Grade 4 glioma, also known as glioblastoma, is studied in a separate TCGA study.

### 4.3 *Melanoma*

The next application of PKB is to a TCGA cutaneous melanoma dataset (TCGA *and others*, 2015). Melanoma is most often discovered after it has metastasized, and the skin melanoma site is never found. Therefore, the majority of the samples are metastatic. In this data set, there are 369 metastatic samples and 103 primary samples. It is of great interest to study the genomic differences between the two types, thus we applied PKB to this data using metastatic/primary as the outcome variable. Using the Biocarta pathways and rbf kernel produced the smallest classification error rate (8.1%) among all methods. Fifteen pathways that PKB found most relevant to the outcome are presented in the third column of Table 4.

Two complement pathways, lectin induced complement pathway and classical complement pathway, came out from the PKB model as the most significant pathways. Proteins in complement system participate in a variety of biological processes of metastasis, such as epithelial-mesenchymal transition (EMT). EMT is an important process in the initiation stage of metastasis, through which cells in primary tumor lose cell-cell adhesion and gain invasive properties. Complement activation by tumor cells can recruit stromal cells to the tumor and induce EMT. Furthermore, complement proteins can mediate the degradation of extracellular matrix, thereby promoting tumor metastasis (Pio *and others*, 2014).

|    | Metabric (grade) | Glioma (grade) | Melanoma (met) |
|----|------------------|----------------|----------------|
| 1  | Cell aggregation | Homophilic cell adhesion via plasma membrane adhesion molecules | Lectin induced complement pathway |
| 2  | Sequestering of metal ion | Neuropeptide signaling pathway | Classical complement pathway |
| 3  | Glutathione derivative metabolic process | Multicellular organismal macromolecule metabolic process | Phospholipase c delta in phospholipid associated cell signaling |
| 4  | Antigen processing and presentation of exogenous peptide antigen via mhc class i | Peripheral nervous system neuron differentiation | Fc epsilon receptor i signaling in mast cells |
| 5  | Sterol biosynthetic process | Positive regulation of hair cycle | Inhibition of matrix metalloproteinases |
| 6  | Pyrimidine containing compound salvage | Peptide hormone processing | Regulation of map kinase pathways through dual specificity phosphatases |
| 7  | Protein dephosphorylation | Hyaluronan metabolic process | Estrogen responsive protein efp controls cell cycle and breast tumors growth |
| 8  | Homophilic cell adhesion via plasma membrane adhesion molecules | Positive regulation of synapse maturation | Chaperones modulate interferon signaling pathway |
| 9  | Cyclooxygenase pathway | Stabilization of membrane potential | Il-10 anti-inflammatory signaling pathway |
| 10 | Establishment of protein localization to endoplasmic reticulum | Lymphocyte chemotaxis | Reversal of insulin resistance by leptin |
| 11 | Negative regulation of dephosphorylation | Insulin secretion | Bone remodeling |
| 12 | Xenophagy | Positive regulation of osteoblast proliferation | Cycling of ran in nucleocytoplasmic transport |
| 13 | Attachment of spindle microtubules to kinetochore | Negative regulation of dephosphorylation | Alternative complement pathway |
| 14 | Fatty acyl coa metabolic process | Trophoblast giant cell differentiation | Cell cycle: g2/m checkpoint |
| 15 | Apical junction assembly | Synaptonemal complex organization | Hop pathway in cardiac development |

Table 4. Top fifteen pathways with the largest weights fitted by PKB. In each column, pathways are sorted in descending order from top to bottom. Pathways in the first two columns are from GO Biological Process pathways, and the third column from Biocarta.

## 5. Discussion

In this paper, we have introduced the PKB model as a method to perform classification analysis of gene expression data, as well as identify pathways relevant to the clinical outcomes of interest. PKB usually yields sparse models in terms of the number of pathways, which enhances interpretability of the results. Moreover, the pathway weights as defined in section 2 can be used as a measure of pathway importance, and provides guidance for further experimental verifications.

Two types of regularizations are introduced in the optimization step of PKB, in order to select simple model with good fitting. The $L_1$ method enjoys a computational advantage due to the sparsity of its solution. In simulations and real data applications, both methods yielded comparable prediction accuracy. It is worth mentioning that the second-order approximation of the log loss function is also necessary for efficiency of PKB. The approximation yields an expression that is quadratic in terms of coefficients $\beta$, which allows the problem to be converted to LASSO or Ridge Regression after regularizations are added. If the original loss function were used, solving $\beta$ would be more time consuming.

There are several limitations of the current PKB approach. First of all, when constructing base learners from pathways, we use fixed bandwidth parameters (inverse of the number of genes in each pathway) in the kernel functions. Ideally, we would like the model to auto-determine the parameters. However, the number of such parameters is equal to the number of pathways, which is often too large to tune efficiently. Therefore, it remains a challenging task for future research. Second, we currently only use pathway as a criterion to group genes, and within each pathway, all genes are treated equally. It is conceivable that the genes interact with each other through an underlying interaction network, and intuitively, genes in the hub should get more weights compared to genes on the periphery. With the network information available, it is possible to build more sensible kernel functions as base learners. Third, the pathway databases only cover a subset of the input genes. Both KEGG and Biocarta only include a few thousands of genes, while the number of input genes is usually beyond 15,000. Large number of genes, with the potential to provide additional prediction power, remain unused in the model. In our applications, we tried pooling together all unused genes and consider them as a new pathway, but it did not significantly improve the results. Although genes annotated with pathways are supposed to be most informative, it is still worth looking for smarter ways of handling unannotated genes.

## 6. Software

Python code along with a complete instruction for usage of PKB is available from website: https://github.com/zengliX/PKB.

## 7. Supplementary Material

Supplementary material and reproduction code are available online at https://github.com/zengliX/PKB. Reproduction-related input data sets are available upon request from the corresponding author (hongyu.zhao@yale.edu).

## Acknowledgments

Acknowledgement section.

*Conflict of Interest*: None declared.

## References

Aiolli, Fabio and Donini, Michele. (2015). Easymkl: a scalable multiple kernel learning algorithm. *Neurocomputing* **169**, 215–224.

Breiman, Leo. (2001). Random forests. *Machine learning* **45**(1), 5–32.

Carlson, Christopher S, Eberle, Michael A, Kruglyak, Leonid and Nickerson,

DEBORAH A. (2004). Mapping complex disease loci in whole-genome association studies. *Nature* **429**(6990), 446.

COSTELLO, JAMES C, HEISER, LAURA M, GEORGII, ELISABETH, GÖNEN, MEHMET, MENDEN, MICHAEL P, WANG, NICHOLAS J, BANSAL, MUKESH, HINTSANEN, PETTERI, KHAN, SULEIMAN A, MPINDI, JOHN-PATRICK *and others*. (2014). A community effort to assess and improve drug sensitivity prediction algorithms. *Nature biotechnology* **32**(12), 1202–1212.

DRUCKER, HARRIS, BURGES, CHRISTOPHER JC, KAUFMAN, LINDA, SMOLA, ALEX J AND VAPNIK, VLADIMIR. (1997). Support vector regression machines. In: *Advances in neural information processing systems*. pp. 155–161.

FRIEDMAN, JEROME, HASTIE, TREVOR AND TIBSHIRANI, ROBERT. (2001). *The elements of statistical learning*, Volume 1. Springer series in statistics New York.

FRIEDMAN, JEROME, HASTIE, TREVOR, TIBSHIRANI, ROBERT *and others*. (2000). Additive logistic regression: a statistical view of boosting (with discussion and a rejoinder by the authors). *The annals of statistics* **28**(2), 337–407.

FRIEDMAN, JEROME H. (2001). Greedy function approximation: a gradient boosting machine. *Annals of statistics*, 1189–1232.

FUKUMIZU, KENJI, BACH, FRANCIS R AND JORDAN, MICHAEL I. (2009). Kernel dimension reduction in regression. *The Annals of Statistics*, 1871–1905.

GÖNEN, MEHMET AND MARGOLIN, ADAM A. (2014). Drug susceptibility prediction against a panel of drugs using kernelized bayesian multitask learning. *Bioinformatics* **30**(17), i556–i563.

GUYON, ISABELLE, WESTON, JASON, BARNHILL, STEPHEN AND VAPNIK, VLADIMIR. (2002). Gene selection for cancer classification using support vector machines. *Machine learning* **46**(1), 389–422.

Hermani, Alexander, Hess, Jochen, De Servi, Barbara, Medunjanin, Senad, Grobholz, Rainer, Trojan, Lutz, Angel, Peter and Mayer, Doris. (2005). Calcium-binding proteins s100a8 and s100a9 as novel diagnostic markers in human prostate cancer. *Clinical Cancer Research* **11**(14), 5146–5152.

Johnson, Rie and Zhang, Tong. (2014). Learning nonlinear functions using regularized greedy forest. *IEEE transactions on pattern analysis and machine intelligence* **36**(5), 942–954.

Kanehisa, Minoru and Goto, Susumu. (2000). Kegg: kyoto encyclopedia of genes and genomes. *Nucleic acids research* **28**(1), 27–30.

Leung, N, Turbide, C, Olson, M, Marcus, V, Jothy, S and Beauchemin, N. (2006). Deletion of the carcinoembryonic antigen-related cell adhesion molecule 1 (ceacam1) gene contributes to colon tumor progression in a murine model of carcinogenesis. *Oncogene* **25**(40), 5527.

Liu, Dawei, Lin, Xihong and Ghosh, Debashis. (2007). Semiparametric regression of multidimensional genetic pathway data: Least-squares kernel machines and linear mixed models. *Biometrics* **63**(4), 1079–1088.

Luan, Yihui and Li, Hongzhe. (2007). Group additive regression models for genomic data analysis. *Biostatistics* **9**(1), 100–113.

Monteith, Gregory R, McAndrew, Damara, Faddy, Helen M and Roberts-Thomson, Sarah J. (2007). Calcium and cancer: targeting ca2+ transport. *Nature reviews. Cancer* **7**(7), 519.

Nishimura, Darryl. (2001). Biocarta. *Biotech Software & Internet Report: The Computer Software Journal for Scient* **2**(3), 117–120.

*REFERENCES* 23Pereira, Bernard, Chin, Suet-Feung, Rueda, Oscar M, Vollan, Hans-Kristian Moen, Provenzano, Elena, Bardwell, Helen A, Pugh, Michelle, Jones, Linda, Russell, Roslin, Sammut, Stephen-John *and others*. (2016). The somatic mutation profiles of 2,433 breast cancers refines their genomic and transcriptomic landscapes. *Nature communications* **7**.

Pio, Ruben, Corrales, Leticia and Lambris, John D. (2014). The role of complement in tumor growth. In: *Tumor microenvironment and cellular stress*. Springer, pp. 229–262.

Schaefer, Carl F, Anthony, Kira, Krupa, Shiva, Buchoff, Jeffrey, Day, Matthew, Hannay, Timo and Buetow, Kenneth H. (2008). Pid: the pathway interaction database. *Nucleic acids research* **37**(suppl_1), D674–D679.

Subramanian, Aravind, Tamayo, Pablo, Mootha, Vamsi K, Mukherjee, Sayan, Ebert, Benjamin L, Gillette, Michael A, Paulovich, Amanda, Pomeroy, Scott L, Golub, Todd R, Lander, Eric S *and others*. (2005). Gene set enrichment analysis: a knowledge-based approach for interpreting genome-wide expression profiles. *Proceedings of the National Academy of Sciences* **102**(43), 15545–15550.

TCGA *and others*. (2015a). Comprehensive, integrative genomic analysis of diffuse lower-grade gliomas. *N Engl J Med* **2015**(372), 2481–2498.

TCGA *and others*. (2015b). Genomic classification of cutaneous melanoma. *Cell* **161**(7), 1681–1696.

Tilan, Jason and Kitlinska, Joanna. (2016). Neuropeptide y (npy) in tumor growth and progression: Lessons learned from pediatric oncology. *Neuropeptides* **55**, 55–66.

Wei, Zhi and Li, Hongzhe. (2007). Nonparametric pathway-based regression models for analysis of genomic data. *Biostatistics* **8**(2), 265–284.

Wu, Michael C, Lee, Seunggeun, Cai, Tianxi, Li, Yun, Boehnke, Michael and Lin, Xihong. (2011). Rare-variant association testing for sequencing data with the sequence kernel association test. *The American Journal of Human Genetics* **89**(1), 82–93.

Zhang, Xing, Xu, Li-hua and Yu, Qiang. (2010). Cell aggregation induces phosphorylation of pecam-1 and pyk2 and promotes tumor cell anchorage-independent growth. *Molecular cancer* **9**(1), 7.